\documentclass[11pt,a4paper]{article}
\usepackage[hyperref]{emnlp2018}
\usepackage{times}
\usepackage{latexsym}
\usepackage{linguex}
\usepackage{leipzig}
\usepackage{cjhebrew}
\usepackage{ucs}
\usepackage[T1]{fontenc}
\usepackage[utf8]{inputenc}
 \usepackage{array,multirow,graphicx}
\usepackage{stackrel}
\usepackage{textcomp}

\usepackage{url}

\aclfinalcopy % Uncomment this line for the final submission
\title{A Characterwise Windowed Approach to Hebrew Morphological Segmentation}

\author{Amir Zeldes \\
  Department of Linguistics \\
  Georgetown University \\
  {\tt amir.zeldes@georgetown.edu}} 

\date{}

\begin{document}
\maketitle
\begin{abstract}
  This paper presents a novel approach to the segmentation of orthographic word 
  forms in contemporary Hebrew, focusing purely on splitting without 
  carrying out morphological analysis or disambiguation. Casting the analysis task 
  as character-wise binary classification and using adjacent character and word-based   
  lexicon-lookup features, this approach achieves over 98\% accuracy on
  the benchmark SPMRL shared task data for Hebrew, and 97\% accuracy on a 
  new out of domain Wikipedia dataset, an improvement of $\approx$4\% and 5\% over previous state of the art performance.  
\end{abstract}

\section{Introduction}

Hebrew is a morphologically rich language in which, as in Arabic and other similar languages, space-delimited word forms contain multiple units corresponding to what other languages (and most POS tagging/parsing schemes) consider multiple words. This includes nouns fused with prepositions, articles and possessive pronouns, as in \ref{cnoun}, or verbs fused with preceding conjunctions and object pronouns, as in \ref{cverb}, which uses one `word' to mean ``and that they found him'', where the first two letters correspond to `and' and `that' respectively, while the last two letters after the verb mean `him'.\footnote{In all Hebrew examples below, angle brackets denote transliterated graphemes with dots separating morphemes, and square brackets represent standard Hebrew pronunciation. Dot-separated glosses correspond to the segments in the transliteration.}

\exg. \begin{cjhebrew}tybhm|\end{cjhebrew}  \\
    $\langle$m.h.byt$\rangle$ $[$me.ha.bayit$]$\\
	\glt  from.the.house\label{cnoun}

\exg. \begin{cjhebrew}whw'.sm/sw\end{cjhebrew}  \\
    $\langle$w.š.mc'w.hw$\rangle$ $[$ve.še.mtsa'u.hu$]$\\
	\glt  and.that.found-3-PL.him\label{cverb}

This complexity makes tokenization for Hebrew a challenging process, usually carried out in two steps: tokenization of large units, delimited by spaces or punctuation as in English, and a subsequent segmentation of the smaller units within each large unit. To avoid confusion, the rest of this paper will use the term `super-token' for large units such as the entire contents of \ref{cnoun} and `sub-token' for the smaller units.

Beyond this, three issues complicate Hebrew segmentation further. Firstly, much like Arabic and similar languages, Hebrew uses a consonantal script which disregards most vowels. For example in \ref{cverb} the sub-tokens \textit{ve} `and' and \textit{še} `that' are spelled as single letters $\langle$w$\rangle$ and $\langle$š$\rangle$. While some vowels are represented (mainly word-final or etymologically long vowels), these are denoted by the same letters as consonants, e.g.~the final -u in \textit{vešemtsa'uhu}, spelled as $\langle$w$\rangle$, the same letter used for \textit{ve} `and'. This means that syllable structure, a crucial cue for segmentation, is obscured.

A second problem is that unlike Arabic, Hebrew has morpheme reductions creating segments which are pronounced, but not distinguished in the orthography.\footnote{A reviewer has mentioned that Arabic does have some reductions, e.g.~in the fusion of articles with the preposition \textit{li-} `to' in \textit{li.l-} `to the'. This is certainly true and problematic for morphological segmentation of Arabic, however in these cases there is always some graphical trace of the existence of the article, whereas in the Hebrew cases discussed here, no trace of the article is found.} This affects the definite article after some prepositions, as in \ref{bebayit} and \ref{babayit}, which mean `in a house' and `in the house' respectively.

\exg. \begin{cjhebrew}tybb\end{cjhebrew}  \\
    $\langle$b.byt$\rangle$ $[$be.bayit$]$\\
	\glt  in.house\label{bebayit}

\exg. \begin{cjhebrew}tybb\end{cjhebrew}  \\
    $\langle$b..byt$\rangle$ $[$b.a.bayit$]$\\
	\glt  in.the.house\label{babayit}

In \ref{babayit}, the definite article \textit{ha} merges with the preposition \textit{be} to produce the pronounced form \textit{ba}; however the lack of vowel orthography means that both forms are spelled alike.

A final problem is the high degree of ambiguity in written Hebrew, which has often been exemplified by the following example, reproduced from \citet{AdlerElhadad2006}, which has a large number of possible analyses.

\exg. \begin{cjhebrew}Ml.sb\end{cjhebrew}  \\
  $\langle$b.cl.m$\rangle$ be.cil.am - in.shadow.their\\
  $\langle$b.clm$\rangle$ (be./b.a.)celem - in.(a/the).image\\
  $\langle$b.clm$\rangle$ (be./b.a.)calam {\small in.(a/the).photographer}\\
  $\langle$bcl.m$\rangle$ bcal.am - onion.their\\
  $\langle$bclm$\rangle$ becelem - Betzelem (organization)\label{bclm}

Because some options are likelier than others, information about possible segmentations, frequencies and surrounding words is crucial. We also note that although there are 7 distinct analyses in \ref{bclm}, in terms of segmenting the orthography, only two positions require a decision: either $\langle$b$\rangle$ is followed by a boundary or not, and either $\langle$l$\rangle$ is or not.

Unlike previous approaches which attempt a complete morphological analysis, in this paper the focus is on pure orthographic segmentation: deciding which characters should be followed by a boundary. Although it is clear that full morphological disambiguation is important, I will argue that a pure splitting task for Hebrew can be valuable if it produces substantially more accurate segmentations, which may be sufficient for some downstream applications (e.g.~sequence-to-sequence MT)\footnote{cf.~\citet{HabashSadat2006} on consequences of pure tokenization for Arabic MT.}, but may also be fed into a morphological disambiguator. As will be shown here, this simpler task allows us to achieve very high accuracy on shared task data, substantially outperforming the pure segmentation accuracy of the previous state of the art while remaining robust against out-of-vocabulary (OOV) items and domain changes, a crucial weakness of existing tools. 

The contributions of this work are threefold: 

\begin{enumerate}
\item A robust, cross-domain state of the art model for pure Hebrew word segmentation, independent of morphological disambiguation, with an open source implementation and pretrained models
\item Introducing and evaluating a combination of shallow string-based features and ambiguous lexicon lookup features in a windowed approach to binary boundary classification
\item Providing new resources: a freely available, out-of-domain dataset from Wikipedia for evaluating pure segmentation; a converted version in the same format as the original Hebrew Treebank data used in previous work; and an expanded morphological lexicon based on Universal POS tags.
\end{enumerate}

\section{Previous Work}

Earlier approaches to Hebrew morphological segmentation include finite-state analyzers \cite{YonaWintner2005} and multiple classifiers for morphological properties feedings into a disambiguation step \cite{ShachamWintner2007}. Lattice based approaches have been used with variants of HMMs and the Viterbi algorithm \cite{AdlerElhadad2006} in order to generate all possible analyses supported by a broad coverage lexicon, and then disambiguate in a further step. The main difficulty encountered in all these approaches is the presence of items missing from the lexicon, either due to OOV lexical items or spelling variation, resulting in missing options for the disambiguator. Additionally, there have been issues in comparing results with different formats, datasets, and segmentation targets, especially before the creation of a standard shared task dataset (see below).

Differently from work on Arabic segmentation, where  state of the art work operates as either a sequence tagging task (often using BIO-like encoding and CRF/RNN sequence models, \citealt{MonroeGreenManning2014}), or a holistic character-wise segmentation ranking task \cite{AbdelaliDarwishDurraniEtAl2016}, work on Hebrew segmentation has focused on segmentation in the context of complete morphological analysis, including categorical feature output (gender, tense, etc.). This is in part due to tasks requiring the reconstruction of orthographically unexpressed articles as in \ref{babayit} and other orthographic changes which require morphological disambiguation, such as recovering base forms of inflected nouns and verbs, and inserting segments which cannot be aligned to any orthographic material, as in \ref{inserted_shel} below. In this example, an unexpressed possessive preposition \textit{šel} is inserted, in contrast to segmentation practices for the same construction e.g.~in Arabic and other languages, where the pronoun sub-token is interpreted as inherently possessive.

\exg. \begin{cjhebrew}htyb\end{cjhebrew}  \\
    $\langle$byt.h$\rangle$ $[$beyt.a$]$ `her house' (house.her)\\
	Target analysis: \\
    $\langle$byt šl hy'$\rangle$ $[$bayit šel hi$]$ `house of she'\label{inserted_shel}

Most recently, the shared task on Statistical Parsing of Morphologically Rich Languages (SPMRL 2013-2015, see \citealt{SeddahKueblerTsarfaty2014}) introduced a standard benchmark dataset for Hebrew morphological segmentation based on the Hebrew Treebank (\citealt{SimaanItaiWinterEtAl2001}; the data has $\approx$94K super-tokens for training and $\approx$21K for testing and development), with a corresponding shared task scored in several scenarios. Systems were scored on segmentation, morphological analysis and syntactic parsing, while working from raw text, from gold segmentation (for morphological analysis and parsing), or from gold segmentation and morphology (for parsing). 

The current state of the art is the open source system \textit{yap} (`yet another parser', \citealt{MoreTsarfaty2016}), based on joint morphological and syntactic disambiguation in a lattice-based parsing approach with a rich morphological lexicon. This will be the main system for comparison in Section \ref{sect_res}. Although yap was designed for a fundamentally different task than the current system, i.e.~canonical rather than surface segmentation (cf.~\citealt{CotterellVieiraSchuetze2016}), it is nevertheless currently also the best performing Hebrew segmenter overall and widely used, making it a good point for comparison. Conversely, while the approach in this paper cannot address morphological disambiguation as \textit{yap} and similar systems do, it will be shown that substantially better segmentation accuracy can be achieved using local features, evaluated character-wise, which are much more robustly attested in the limited training data available.

\section{Character-wise Classification}

In this paper we cast the segmentation task as character-wise binary classification, similarly to approaches in other languages, such as Chinese \cite{LeeHuang2013}. The goal is to predict for each character in a super-token (except the last) whether it should be followed by a boundary. Although the character-wise setup prevents a truly global analysis of word formation in the way that a lattice-based model allows, it turns out to be more robust than previous approaches, in large part because it does not require a coherent full analysis in the case of OOV items (see Section \ref{sect_res}).

\subsection{Feature Extraction}\label{feats}

Our features target a window of \textit{n} characters around the character being considered for a following boundary (the `target' character), as well as characters in the preceding and following super-tokens. In practice we set \textit{n} to $\pm 2$, meaning each classification mainly includes information about the preceding and following two characters. 

The extracted features for each window can be divided into three categories: character identity features (i.e.~which letters are observed), numerical position/length features and lexicon lookup features. The lexicon is based on the fully inflected form lexicon created by the MILA center, also used by \citet{MoreTsarfaty2016}, but POS tags have been converted into much less sparse Universal POS tags (cf.~\citealt{PetrovDasMcDonald2012}) matching the Hebrew Treebank data set. Entries not corresponding to tags in the data (e.g.~compound items that would not be a single unit, or additional POS categories) receive the tag X, while complex entries (e.g.~NOUN + possessive) receive a tag affixed with `CPLX'. Multiple entries per word form are possible (see Section \ref{lex_ext}).

\textbf{Character features} For each target character and the surrounding 2 characters in either direction, we consider the identity of the letter or punctuation sign from the set \{\textquotedbl,-,\%,\textquotesingle,.,?,!,/\} as a categorical feature (using native Hebrew characters, not transliteration). Unavailable characters (less than 2 characters from beginning/end of word) and OOV characters are substituted by the underscore. We also encode the first and last letter of the preceding and following super-tokens in the same way. The rationale for choosing these letters is that any super-token always has a first and last character, and in Hebrew these will also encode definite article congruence and clitic pronouns in adjacent super-tokens, which are important for segmenting units. Finally, for each of the five character positions within the target super-token itself, we also encode a boolean feature indicating whether the character could be `vocalic', i.e.~whether it is one of the letters sometimes used to represent a vowel: \{',h,w,y\}. Though this is ostensibly redundant with letter identity, it is actually helpful in the final model (see Section \ref{ablations}).

\textbf{Lexicon lookup} Lexicon lookup is performed for several substring ranges, chosen using the development data: the entire super-token, and characters up to the target inclusive and exclusive; the entire remaining super-token characters inclusive and exclusive; the remaining substrings starting at target positions -1 and -2; 1, 2, 3 and 4-grams around the window including the target character; and the entire preceding and following super-token strings. Table \ref{lookup_feats} illustrates this for character \#3 in the middle super-token from the trigram in \ref{lookup_ex}. The value of the lexicon lookup is a categorical variable consisting of all POS tags attested for the substring in the lexicon, concatenated with a separator into a single string. 

\exg. \begin{cjhebrew}'wh\end{cjhebrew}\quad\begin{cjhebrew}ynkphm/s\end{cjhebrew}\quad\begin{cjhebrew}wnb/sx\end{cjhebrew} \\
    $\langle$xšbnw\quad\quad\quad{}š.mhpkny\quad\quad\quad\quad ~~hw'$\rangle$ \\
    $[$xašavnu\quad\quad še.mahapxani\quad\quad ~hu$]$\\
    thought-1PL that.revolutionary\quad{}is \\
    `we thought that it is revolutionary'\label{lookup_ex}

\begin{table}[t!]
\begin{center}
\begin{tabular}{|l|ll|}
\hline \bf \bf location & \bf substring & \bf lexicon response \\ \hline

 super token & [šmhpkny] & \_ \\
 str so far & [šmh]... & \small ADV$|$NOUN$|$VERB \\
 str remaining & ..[pkny] & \_ \\
 str -1 remain & ..[hpkny] & \_ \\
 str -2 remain & .[mhpkny] & \small ADJ$|$NOUN$|$CPLXN \\
 str from -4 & [\_\_šmh].... & \small \_ \\
 str from -3 & [\_šmh].... & \small \_ \\
 str from -2 & [šmh].... & \small ADV$|$NOUN$|$VERB \\
 str from -1 & .[mh].... & \small ADP$|$ADV \\
 str to +1 & ..[hp]... & \_ \\
 str to +2 & ..[hpk].. & \small NOUN$|$VERB \\
 str to +3 & ..[hpkn]. & \_ \\
 str to +4 & ..[hpkny] & \_ \\
prev string & [xšbnw] & \small VERB \\
next string & [hw'] & \small PRON$|$COP \\

\hline
\end{tabular}
\end{center}
\caption{\label{lookup_feats} Lexicon lookup features for character 3 in the super-token \textit{š.mhpkny}. Overflow positions (e.g.~substring from char -4 for the third character) return `\_'.}
\end{table}

\textbf{Numerical features} To provide information about super-token shape independent of lexicon lookup, we also extract the super-token lengths of the current, previous and next super-token, and encode the numerical position of the character being classified as integers. We also experimented with a `frequency' feature, representing the ratio of the product of frequencies of the substrings left and right of a split under consideration divided by the frequency of the whole super-token. Frequencies were taken from the IsraBlog dataset word counts provided by \citet{Linzen2009}.\footnote{Online at \url{http://tallinzen.net/frequency/}} While this fraction in no way represents the probability of a split, it does provide some information about the relative frequency of parts versus whole in a naive two-segment scenario, which can occasionally help the classifier decide whether to segment ambiguous cases (although this feature's contribution is small, it was retained in the final model, see Section \ref{ablations}).

\textbf{Word embeddings} For one of the learning approaches tested, a deep neural network (DNN) classifier, word embeddings from a Hebrew dump of Wikipedia were added for the current, preceding and following super-tokens, which were expected to help in identifying the kinds of super-tokens seen in OOV cases. However somewhat surprisingly, the approach using these features turned out not to deliver the best results despite outperforming the previous state of the art, and these features are not used in the final system (see Section \ref{sect_res}).

Note that although we do not encode word identities for any super-token, and even less so for models not using word embeddings, length information in conjunction with lexicon lookup and first and last character can already give a strong indication of the surrounding context, at least for those words which turn out to be worth learning. For example, the frequent purpose conjunction $\langle$kdy$\rangle$ \textit{kedey} `in order to', which strongly signals a following infinitive whose leading $\langle$l$\rangle$ should not be segmented, is uniquely identifiable as a three letter conjunction (tag SCONJ) beginning with $\langle$k$\rangle$ and ending with $\langle$y$\rangle$. This combination, if relevant for segmentation, can be learned and assigned weights for each of the characters in adjacent words.

\subsection{Learning Approach}\label{learning}

Several learning approaches were tested for the boundary classification task, including decision tree based classifiers, such as a Random Forest classifier, the Extra Trees variant of the same algorithm \cite{GeurtsErnstWehenkel2006}, and Gradient Boosting, all using the implementation in scikit-learn \cite{PedregosaVaroquauxGramfortEtAl2011}, as well as a DNN classifier implemented in TensorFlow \cite{AbadiBarhamChenEtAl2016}. An initial attempt using sequence-to-sequence learning with an RNN was abandoned early as it underperformed other approaches, possibly due to the limited size of the training data.

For tree-based classifiers, letters and categorical lexicon lookup values were pseudo-ordinalized into integers (using a non-meaningful alphabetic ordering), and numerical features were retained as-is. For the DNN, the best feature representation based on the validation set was to encode characters and positions as one-hot vectors, lexicon lookup features as trainable dense embeddings, and to bucketize length features in single digit buckets up to a maximum of 15, above which all values are grouped together. Additionally, we prohibit segmentations between Latin letters and digits (using regular expressions), and forbid producing any prefix/suffix not attested in training data, ruling out rare spurious segmentations.

\section{Experimental Setup}

\subsection{Data}

\begin{figure*}[hbt]
\centering
\fbox{
\includegraphics[width=\textwidth, trim={2.5cm 20.2cm 2.5cm 2.1cm}, clip]{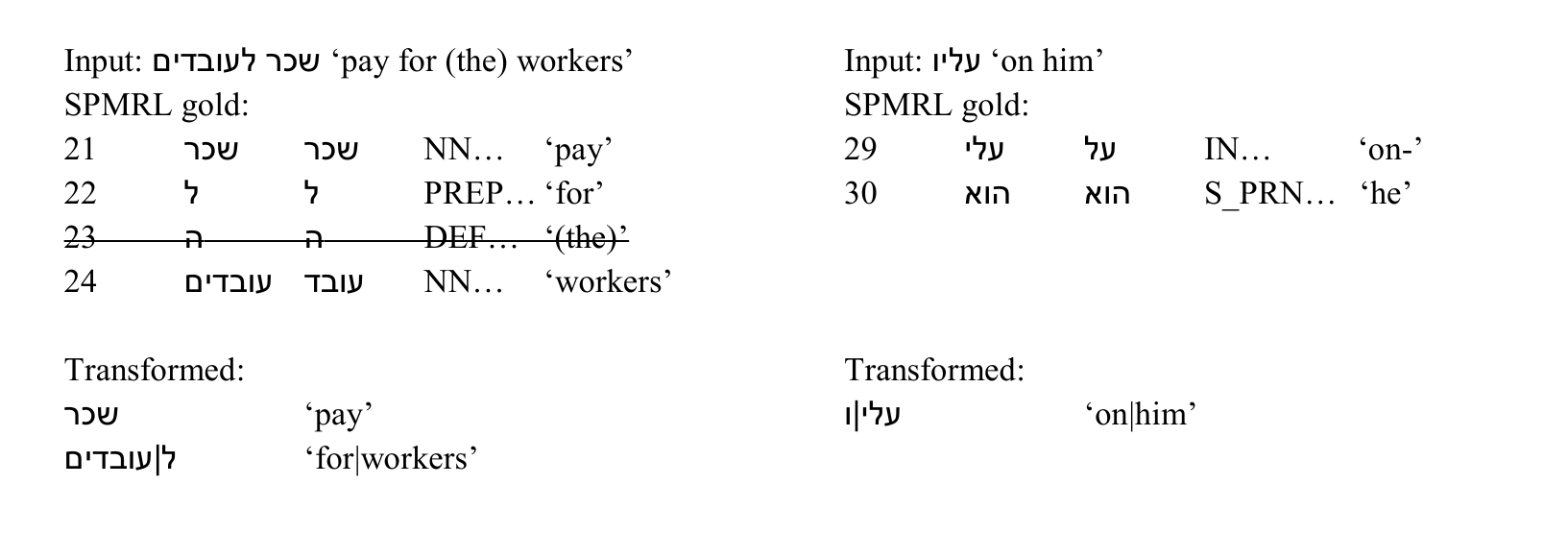}}
\caption{Transformation of SPMRL style data to pure segmentation information. Left: inserted article is deleted; right: a clitic pronoun form is restored.}
\label{SPMRL_norm}
\end{figure*}

Evaluation was done on two datasets. The benchmark SPMRL dataset was transformed by discarding all inserted sub-tokens not present in the input super-tokens (i.e.~inserted articles). We revert alterations due to morphological analysis (de-inflection to base forms), and identify the segmentation point of all clitic pronouns, prepositions etc., marking them in the super-tokens.\footnote{For comparability with previous results, we use the exact dataset and splits used by \citet{MoreTsarfaty2016}, despite a known mid-document split issue that was corrected in version 2 of the Hebrew Treebank data. We thank the authors for providing the data and for help reproducing their setup.} Figure \ref{SPMRL_norm} illustrates the procedure, which was manually checked and automatically verified to reconstitute correctly back into the input data. 

A second test set of 5,000 super-tokens (7,100 sub-tokens) was constructed from another domain to give a realistic idea of performance outside the training domain. While SPMRL data is taken from newswire material with highly standard orthography, this dataset was taken from Hebrew Wikipedia NER data made available by \citet{Al-RfouKulkarniPerozziEtAl2015}. Since that dataset was not segmented into subtokens, manual segmentation of the first 5,000 tokens was carried out, which represent shuffled sentences from a wide range of topics. This data set is referred to below as `Wiki5K'. Both datasets are provided along with the code for this paper via GitHub\footnote{\url{https://github.com/amir-zeldes/RFTokenizer}} as new test sets for future evaluations of Hebrew segmentation.

\subsection{Lexicon extension}\label{lex_ext}

A major problem in Hebrew word segmentation is dealing with OOV items, and especially those due not to regular morphological processes, but to foreign names. If a foreign name begins with a letter that can be construed as a prefix, and neither the full name nor the substring after the prefix is attested in the lexicon, the system must resort to purely contextual information for disambiguation. As a result, having a very complete list of possible foreign names is crucial. 

The lexicon mentioned above has very extensive coverage of native items as well as many foreign items, and after tag-wise conversion to Universal POS tags, contains over 767,000 items, including multiple entries for the same string with different tags. However its coverage of foreign names is still partial. In order to give the system access to a broader range of foreign names, we expanded the lexicon with proper name data from three sources:

\begin{itemize}
  \item WikiData persons with a Hebrew label, excluding names whose English labels contain determiners, prepositions or pronouns
  \item WikiData cities with a Hebrew label, again excluding items with determiners, prepositions or pronouns in English labels
  \item All named entities from the Wikipedia NER data found later than the initial 5K tokens used for the Wiki5K data set
\end{itemize}

These data sources were then white-space tokenized, and all items which could spuriously be interpreted as a Hebrew article/preposition + noun were removed. For example, a name `Leal' $\langle$l'l$\rangle$ is excluded, since it can interfere with segmenting the sequence $\langle$l.'l$\rangle$ \textit{la'el} `to God'. This process added over 15,000 items, or $\approx$2\% of lexicon volume, all labeled PROPN. In Section \ref{ablations} performance with and without this extension is compared.

\subsection{Evaluation Setup}

Since comparable segmentation systems, including the previous state of the art, do insert reconstructed articles and carry out base form transformations (i.e.~they aim to produce the gold format in Figure \ref{SPMRL_norm}), we do not report or compare results to previously published scores. All systems were re-run on both datasets, and no errors were counted where these resulted from data alterations. Specifically, other systems are never penalized for reconstructing or omitting inserted units such as articles, and when constructing clitics or alternate base forms of nouns or verbs, only the presence or absence of a split was checked, not whether inferred noun or verb lemmas are correct. This leads to higher scores than previously reported, but makes results comparable across systems on the pure segmentation task.

\section{Results}\label{sect_res}

Table \ref{performance} shows the results of several systems on both datasets.\footnote{An anonymous reviewer suggested that it would also be interesting to add an unsupervised pure segmentation system such as Morfessor \cite{CreutzLagus2002} to evaluate on this task. This would certainly be an interesting comparison, but due to the brief response period it was not possible to add this experiment before publication. It can however be expected that results would be substantially worse than yap, given the centrality of the lexicon representation to this task, which can be seen in detail in the ablation tests in Section \ref{ablations}.} The column `\% perf' indicates the proportion of perfectly segmented super-tokens, while the next three columns indicate precision, recall and F-score for boundary detection, not including the trivial final position characters.

\begin{table}[t!]
\begin{center}
\begin{tabular}{|l|r|r|r|r|}
\hline  & \bf \% perf & \bf P & \bf R & \bf F \\ 
\hline
\bf SPMRL & & & & \\
  \hline
 \textit{baseline} & 69.65 & -- & -- & -- \\
 \textit{UDPipe} & 89.65 & 93.52 & 68.82 & 79.29 \\
 \textit{yap} & 94.25 & 86.33 & 96.33 & 91.05 \\
 \textit{RF} & \bf{98.19} & \bf{97.59} & \bf{96.57} & \bf{97.08} \\
 \textit{DNN} & 97.27 & 95.90 & 95.01 & 95.45 \\
  \hline
\bf Wiki5K & & & & \\
  \hline
 \textit{baseline} & 67.61 & -- & -- & -- \\
 \textit{UDPipe} & 87.39 & 92.03 & 64.88 & 76.11 \\
 \textit{yap} & 92.66 & 85.55 & 92.34 & 88.81 \\
 \textit{RF} & \bf 97.63 & \bf 97.41 & \bf 95.31 & \bf 96.35 \\
 \textit{DNN} & 95.72 & 94.95 & 92.22 & 93.56 \\

\hline

\end{tabular}
\end{center}
\caption{\label{performance} System performance on the SPMRL and Wiki5K datasets.}
\end{table}

The first baseline strategy of not segmenting anything is given in the first row, and unsurprisingly gets many cases right, but performs badly overall. A more intelligent baseline is provided by UDPipe (\citealt{StrakaHajicStrakova2016}; retrained on the SPMRL data), which, for super-tokens in morphologically rich languages such as Hebrew, implements a `most common segmentation' baseline (i.e.~each super-token is given its most common segmentation from training data, forgoing segmentation for OOV items).\footnote{UDPipe also implements an RNN tokenizer to segment punctuation spelled together with super-tokens; however since the evaluation dataset already separates such punctuation symbols, this component can be ignored here.} Results for yap represent pure segmentation performance from the previous state of the art \cite{MoreTsarfaty2016}.

The best two approaches in the present paper are represented next: the Extra Trees Random Forest variant,\footnote{Extra Trees outperformed Gradient Boosting and Random Forest in hyperparameter selection tuned on the dev set. Using a grid search led to the choice of 250 estimators (tuned in increments of 10), with unlimited features and default scikit-learn values for all other parameters.} called RFTokenizer, is labeled  \textit{RF} and the DNN-based system is labeled \textit{DNN}. Surprisingly, while the DNN is a close runner up, the best performance is achieved by the RFTokenizer, despite not having access to word embeddings. Its high performance on the SPMRL dataset makes it difficult to converge to a better solution using the DNN, though it is conceivable that substantially more data, a better feature representation and/or more hyperparameter tuning could equal or surpass the RFTokenizer's performance. Coupled with a lower cost in system resources and external dependencies, and the ability to forgo large model files to store word embeddings, we consider the RFTokenizer solution to be better given the current training data size.

Performance on the out of domain dataset is encouragingly nearly as good as on SPMRL, suggesting our features are robust. This is especially clear compared to UDPipe and yap, which degrade more substantially. A key advantage of the present approach is its comparatively high precision. While other approaches have good recall, and yap approaches RFTokenizer on recall for SPMRL, RFTokenizer's reduction in spurious segmentations boosts its F-score substantially. To see why, we examine some errors in the next section, and perform feature ablations in the following one.

\section{Error Analysis}

Looking at the SPMRL data, RFTokenizer makes relatively few errors, the large majority of which belong to two classes: morphologically ambiguous cases with known vocabulary, as in \ref{kitsa}, and OOV items, as in \ref{varela}. In \ref{kitsa}, the sequence $\langle$qch$\rangle$ could be either the noun \textit{katse} `edge' (single super-token and sub-token), or a possessed noun \textit{kits.a} `end of FEM-SG' with a final clitic pronoun possessor (two sub-tokens). The super-token coincidentally begins a sentence ``The end of a year ...'', meaning that the preceding unit is the relatively uninformative period, leaving little local context for disambiguating the nouns, both of which could be followed by \textit{šel} `of'.

\exg. \begin{cjhebrew}hn/s\end{cjhebrew}~\begin{cjhebrew}l/s\end{cjhebrew} ~\begin{cjhebrew}h.sq\end{cjhebrew} \\
    Gold: \quad\quad\quad\quad\quad\quad\quad~Pred: \\
    $\langle$qc.h šl šnh$\rangle$\quad\quad\quad\quad $\langle$qch šel šnh$\rangle$ \\
    $[$kits.a šel šana$]$ \quad\quad\quad$[$katse šel šana$]$\\
    end.SG-F of year \quad\quad~edge of year \\
    ``The end of a year'' \quad``An edge of a year''\label{kitsa}
       
A typical case of an OOV error can be seen in \ref{varela}. In this example, the lexicon is missing the name $\langle$w'r'lh$\rangle$ `Varela', but does contain a name $\langle$'r'lh$\rangle$ `Arela'. As a result, given the context of a preceding name `Maria', the tokenizer opts to recognize a shorter proper name and assigns the letter `w' to be the word `and'.

\ex. . \begin{cjhebrew}hl'r'w\end{cjhebrew} \begin{cjhebrew}hyrm|\end{cjhebrew}\label{varela} \\
    Gold: \quad\quad\quad\quad\quad Pred: \\
    $\langle$mryh w'r'lh .$\rangle$ \quad $\langle$mryh w.'r'lh .$\rangle$ \\
    $[$mariya varela$]$ \quad $[$mariya ve.arela$]$ \\
    `Maria Varela.'  \quad `Maria and Arela.'

To understand the reasons for RFTokenizer's higher precision compared to other tools, it is useful to consider errors which RFTokenizer succeeds in avoiding, as in \ref{barmore}-\ref{lehitamen} (only a single bold-faced word is discussed in detail for space reasons; broader translations are given for context, keeping Hebrew word order). In \ref{barmore}, RF and yap both split \textit{w} `and' from the OOV item $\langle$bwrmwr$\rangle$ `Barmore' correctly. The next possible boundary, `b.w', is locally unlikely, as a spelling `bw' makes a reading $[$bo$]$ or $[$bu$]$ likely, which is incompatible with the segmentation. However, yap considers global parsing likelihood, and the verb `run into' takes the preposition \textit{b} `in'. It segments the `b' in the OOV item, a decision which RFTokenizer avoids based on low local probabilities. 

\ex. ``ran into Campbell and \textbf{Barmore}'' \label{barmore} \\
    RF: \quad\quad\quad\quad\quad yap: \\
    $\langle$w.bwrmwr$\rangle$ \quad $\langle$w.b.wrmwr$\rangle$

\ex. ``meanwhile continues the player, who returned to practice last week, \textbf{to-train}'' \label{lehitamen} \\
    RF: \quad\quad\quad  yap: \\
    $\langle$lht'mn$\rangle$ \quad $\langle$l.ht'm.n$\rangle$

In \ref{lehitamen}, RF leaves the medium frequency verb `to train' unsegmented. By contrast, yap considers the complex sentence structure and long distance to the fronted verb `continues', and prefers a locally very improbable segmentation into the preposition \textit{l} `to', a noun \textit{ht'm} `congruence' and a 3rd person feminine plural possessive \textit{n}: `to their congruence'. Such an analysis is not likely to be guessed by a native speaker shown this word in isolation, but becomes likelier in the context of evaluating possible parse lattices with limited training data.

We speculate that lower reliance on complete parses makes RFTokenizer more robust to errors, since data for character-wise decisions is densely attested. In some cases, as in \ref{barmore}, it is possible to segment individual characters based on similarity to previously seen contexts, without requiring super-tokens to be segmentable using the lexicon. This is especially important for partially correct results, which affect recall, but not necessarily the percentage of perfectly segmented super-tokens.

In Wiki5K we find more errors, degrading performance $\approx$0.7\%. Domain differences in this data lead not only to OOV items (esp. foreign names), but also distributional and spelling differences. In \ref{hartberg}, heuristic segmentation based on a single character position backfires, and the tokenizer over-zealously segments. This is due to neither the place `Hartberg', nor a hypothetical `Retberg' being found in the lexicon, and the immediate context being uninformative surrounding commas.

\ex. , \begin{cjhebrew}grb.trh\end{cjhebrew} ,\label{hartberg} \\
    Gold: \quad\quad\quad\quad Pred: \\
    $\langle$hr$\stackrel[^\textrm{.}]{}{\textrm{t}}$brg$\rangle$ \quad\quad\quad$\langle$h.r$\stackrel[^\textrm{.}]{}{\textrm{t}}$brg$\rangle$ \\
    $[$hartberg$]$\quad\quad\quad$[$ha.retberg$]$ \\
    `Hartberg'\quad\quad`the Retberg'

\section{Ablation tests}\label{ablations}

Table \ref{ablation_table} gives an overview of the impact on performance when specific features are removed: the entire lexicon, lexicon expansion, letter identity, `vowel' features from Section \ref{feats}, and both of the latter. Performance is high even in ablation scenarios, though we keep in mind that baselines for the task are high (e.g.~`most frequent lookup', the UDPipe strategy, achieves close to 90\%). 

The results show the centrality of the lexicon: removing lexicon lookup features degrades performance by about 3.5\% perfect accuracy, or 5.5 F-score points. All other ablations impact performance by less than 1\% or 1.5 F-score points. Expanding the lexicon using Wikipedia data offers a contribution of 0.3--0.4 points, confirming the original lexicon's incompleteness.\footnote{An anonymous reviewer has asked whether the same resources from the NER dataset have been or could be made available to the competing systems. Unfortunately it was not possible to re-train yap using this data, since the lexicon used by that system has a much more complex structure compared to the simple PROPN tags used in our approach (i.e.~we would need to codify much richer morphological information for the added words). However the ablations  show that even without the expanded lexicon, RFTokenizer outperforms yap by a large margin. For UDPipe no lexicon is used, so that this issue does not arise.}

\begin{table}[t!]
\begin{center}
\begin{tabular}{|l|r|r|r|r|}
\hline  & \bf \% perf & \bf P & \bf R & \bf F \\ 
\hline
\bf SPMRL & & & & \\
  \hline
 FINAL & 98.19 & 97.59 & 96.57 & 97.08 \\
 -expansion & 98.01 & 97.25 & 96.35 & 96.80 \\
 -vowels & 97.99 & 97.55 & 95.97 & 96.75 \\
 -letters & 97.77 & 96.98 & 95.73 & 96.35 \\
 -letr-vowl & 97.57 & 97.56 & 94.44 & 95.97 \\
 -lexicon & 94.79 & 92.08 & 91.46 & 91.77 \\
  \hline
\bf Wiki5K & & & & \\
  \hline
 FINAL & 97.63 & 97.41 & 95.31 & 96.35 \\
 -expansion & 97.33 & 96.64 & 95.31 & 95.97 \\
 -vowels & 97.51 & 97.56 & 94.87 & 96.19 \\
 -letters & 97.27 & 96.89 & 94.71 & 95.79 \\
 -letr-vowl & 96.72 & 97.17 & 92.77 & 94.92 \\
 -lexicon & 94.72 & 92.53 & 91.51 & 92.01  \\
\hline
\end{tabular}
\end{center}
\caption{\label{ablation_table} Effects of removing features on performance, ordered by descending F-score impact on SPMRL.}
\end{table}

Looking more closely at the other features, it is surprising that identity of the letters is not crucial, as long as we have access to dictionary lookup using the letters. Nevertheless, removing letter identity impacts especially boundary recall, perhaps because some letters receive identical lookup values (e.g.~single letter prepositions such as \textit{b} `in', \textit{l} `to') but have different segmentation likelihoods.

The `vowel' features, though ostensibly redundant with letter identity, help a little, causing 0.33 SPMRL F-score point degradation if removed. A cursory inspection of differences with and without vowel features indicates that adding them allows for stronger generalizations in segmenting affixes, especially clitic pronouns (e.g.~if a noun is attested with a `vocalic' clitic like \textit{h} `hers', it generalizes better to unseen cases with \textit{w} `his'). In some cases, the features help identify phonotactically likely splits in a `vowel' rich environment, as in \ref{vowel_err} with the sequence $\langle$hyy$\rangle$ which is segmented correctly in the +vowels setting. 

\exg. \begin{cjhebrew}Nktyyh\end{cjhebrew}  \\
    +Vowels: \quad\quad\quad\quad -Vowels: \\
    $\langle$h.yytkn$\rangle$ \quad\quad\quad $\langle$hyytkn$\rangle$ \\
    $[$ha.yitaxen$]$ \quad\quad\quad \\
    QUEST.possible \\
    `is it possible?'\quad \label{vowel_err}

Removing both letter and vowel features essentially reduces the system to using only the surrounding POS labels. However since classification is character-wise and a variety of common situations can nevertheless be memorized, performance does not break down drastically. The impact on Wiki5k is stronger, possibly because the necessary memorization of familiar contexts is less effective out of domain.

\section{Discussion}

This paper presented a character-wise approach to Hebrew segmentation, relying on a combination of shallow surface features and windowed lexicon lookup features, encoded as categorical variables concatenating possible POS tags for each window. Although the approach does not correspond to a manually created finite state morphology or a parsing-based approach, it can be conjectured that the sequence of possible POS tag combinations at each character position in a sequence of words gives a similar type of information about possible transitions at each potential boundary.

The character-wise approach turned out to be comparatively robust, possibly thanks to the dense training data available, when compared to the smaller order of magnitude if data is interpreted with each super-token, or even each sentence forming a single observation. Nevertheless, there are multiple limitations to the current approach.

Firstly, RFTokenizer does not reconstruct unexpressed articles. Although this is an important task in Hebrew NLP, it can be argued that definiteness annotation can be performed as part of morphological analysis after basic segmentation has been carried out. An advantage of this approach is that the segmented data corresponds perfectly to the input string, reducing processing efforts needed to keep track of the mapping of raw and tokenized data.

Secondly, there is still room for improvement, and it seems surprising that the DNN approach with embeddings could not outperform the RF approach. More training data is likely to make DNN/RNN approaches more effective, similarly to recent advances in tokenization for languages such as Chinese (cf.~\citealt{CaiZhao2016}, though we recognize Hebrew segmentation is much more ambiguous, and embeddings are likely more useful for ideographic scripts).\footnote{During the review period of this paper, a paper by \citet{ShaoHardmeierNivre2018} appeared which nearly matches the performance of yap on Hebrew segmentation using an RNN approach. Achieving an F-score of 91.01 compared to yap's score of 91.05, but on a dataset with slightly different splits, this system gives a good baseline for a tuned RNN-based system. However comparing to RFTokenizer's score of 97.08, it is clear that while RNNs can also do well on the current task, there is still a substantial gap compared to the windowed, lexicon-based binary classification approach take here.} We are currently experimenting with word representations optimized to the segmentation task, including using embeddings or Brown clusters grouping super-tokens with different distributions. Finally, the frequency data obtained from \citet{Linzen2009} is relatively small (only 20K forms), and not error-free due to automatic processing, meaning that extending this data source may yield improvements as well. 

\section*{Acknowledgments}

This work benefited from research on morphological segmentation for Coptic, funded by the National Endowment for the Humanities (NEH) and the German Research Foundation (DFG) (grants HG-229371 and HAA-261271). Thanks are also due to Shuly Wintner, Nathan Schneider and the anonymous reviewers for valuable comments on earlier versions of this paper, as well as to Amir More and Reut Tsarfaty for their help in reproducing the experimental setup for comparing yap and other systems with RFTokenizer.

\bibliography{emnlp2018}

\begin{thebibliography}{21}
\expandafter\ifx\csname natexlab\endcsname\relax\def\natexlab#1{#1}\fi

\bibitem[{Abadi et~al.(2016)Abadi, Barham, Chen, Chen, Davis, Dean, Devin,
  Ghemawat, Irving, Isard, Kudlur, Levenberg, Monga, Moore, Murray, Steiner,
  Tucker, Vasudevan, Warden, Wicke, Yu, and Zheng}]{AbadiBarhamChenEtAl2016}
Mart\'{\i}n Abadi, Paul Barham, Jianmin Chen, Zhifeng Chen, Andy Davis, Jeffrey
  Dean, Matthieu Devin, Sanjay Ghemawat, Geoffrey Irving, Michael Isard,
  Manjunath Kudlur, Josh Levenberg, Rajat Monga, Sherry Moore, Derek~G. Murray,
  Benoit Steiner, Paul Tucker, Vijay Vasudevan, Pete Warden, Martin Wicke, Yuan
  Yu, and Xiaoqiang Zheng. 2016.
\newblock Tensor{F}low: A system for large-scale machine learning.
\newblock In \emph{Proceedings of the 12th USENIX Conference on Operating
  Systems Design and Implementation}, pages 265--283, Savannah, GA.

\bibitem[{Abdelali et~al.(2016)Abdelali, Darwish, Durrani, and
  Mubarak}]{AbdelaliDarwishDurraniEtAl2016}
Ahmed Abdelali, Kareem Darwish, Nadir Durrani, and Hamdy Mubarak. 2016.
\newblock Farasa: A fast and furious segmenter for {A}rabic.
\newblock In \emph{Proceedings of the NAACL 2016: Demonstrations}, pages
  11--16, San Diego, CA.

\bibitem[{Adler and Elhadad(2006)}]{AdlerElhadad2006}
Meni Adler and Michael Elhadad. 2006.
\newblock An unsupervised morpheme-based {HMM} for {H}ebrew morphological
  disambiguation.
\newblock In \emph{Proceedings of the 21st International Conference on
  Computational Linguistics and 44th Annual Meeting of the ACL}, pages
  665--672, Sydney.

\bibitem[{Al-Rfou et~al.(2015)Al-Rfou, Kulkarni, Perozzi, and
  Skiena}]{Al-RfouKulkarniPerozziEtAl2015}
Rami Al-Rfou, Vivek Kulkarni, Bryan Perozzi, and Steven Skiena. 2015.
\newblock Polyglot-{NER}: Massive multilingual named entity recognition.
\newblock In \emph{Proceedings of the 2015 {SIAM} International Conference on
  Data Mining}, Vancouver, Canada.

\bibitem[{Cai and Zhao(2016)}]{CaiZhao2016}
Deng Cai and Hai Zhao. 2016.
\newblock Neural word segmentation learning for {C}hinese.
\newblock In \emph{Proceedings of ACL 2016}, pages 409--420, Berlin.

\bibitem[{Cotterell et~al.(2016)Cotterell, Vieira, and
  Sch\"{u}tze}]{CotterellVieiraSchuetze2016}
Ryan Cotterell, Tim Vieira, and Hinrich Sch\"{u}tze. 2016.
\newblock A joint model of orthography and morphological segmentation.
\newblock In \emph{Proceedings of NAACL-HLT 2016}, pages 664--669, San Diego,
  CA.

\bibitem[{Creutz and Lagus(2002)}]{CreutzLagus2002}
Mathias Creutz and Krista Lagus. 2002.
\newblock Unsupervised discovery of morphemes.
\newblock In \emph{Proceedings of the Workshop on Morphological and
  Phonological Learning at ACL 2002}, pages 21--30, Philadelphia, PA.

\bibitem[{Geurts et~al.(2006)Geurts, Ernst, and
  Wehenkel}]{GeurtsErnstWehenkel2006}
Pierre Geurts, Damien Ernst, and Louis Wehenkel. 2006.
\newblock Extremely randomized trees.
\newblock \emph{Machine Learning}, 63(1):3--42.

\bibitem[{Habash and Sadat(2006)}]{HabashSadat2006}
Nizar Habash and Fatiha Sadat. 2006.
\newblock {A}rabic preprocessing schemes for statistical machine translation.
\newblock In \emph{Proceedings of NAACL 2006}, pages 49--52, New York.

\bibitem[{Lee and Huang(2013)}]{LeeHuang2013}
Chia-Ming Lee and Chien-Kang Huang. 2013.
\newblock Context-based {C}hinese word segmentation using {SVM}
  machine-learning algorithm without dictionary support.
\newblock In \emph{Sixth International Joint Conference on Natural Language
  Processing}, pages 614--622, Nagoya, Japan.

\bibitem[{Linzen(2009)}]{Linzen2009}
Tal Linzen. 2009.
\newblock Corpus of blog postings collected from the {I}srablog website.
\newblock Technical report, Tel Aviv University, Tel Aviv.

\bibitem[{Monroe et~al.(2014)Monroe, Green, and
  Manning}]{MonroeGreenManning2014}
Will Monroe, Spence Green, and Christopher~D. Manning. 2014.
\newblock Word segmentation of informal {A}rabic with domain adaptation.
\newblock In \emph{Proceedings of the 52nd Annual Meeting of the Association
  for Computational Linguistics}, pages 206--211, Baltimore, MD.

\bibitem[{More and Tsarfaty(2016)}]{MoreTsarfaty2016}
Amir More and Reut Tsarfaty. 2016.
\newblock Data-driven morphological analysis and disambiguation for
  morphologically-rich languages and universal dependencies.
\newblock In \emph{Proceedings of {COLING} 2016}, pages 337--348, Osaka, Japan.

\bibitem[{Pedregosa et~al.(2011)Pedregosa, Varoquaux, Gramfort, Michel,
  Thirion, Grisel, Blondel, Prettenhofer, Weiss, and
  Dubourg}]{PedregosaVaroquauxGramfortEtAl2011}
Fabian Pedregosa, Ga{\"e}l Varoquaux, Alexandre Gramfort, Vincent Michel,
  Bertrand Thirion, Olivier Grisel, Mathieu Blondel, Peter Prettenhofer, Ron
  Weiss, and Vincent Dubourg. 2011.
\newblock Scikit-learn: Machine learning in {P}ython.
\newblock \emph{Journal of machine learning research}, 12:2825--2830.

\bibitem[{Petrov et~al.(2012)Petrov, Das, and McDonald}]{PetrovDasMcDonald2012}
Slav Petrov, Dipanjan Das, and Ryan McDonald. 2012.
\newblock A universal part-of-speech tagset.
\newblock In \emph{Proceedings of LREC 2012}, pages 2089--2096, Istanbul,
  Turkey.

\bibitem[{Seddah et~al.(2014)Seddah, K\"{u}bler, and
  Tsarfaty}]{SeddahKueblerTsarfaty2014}
Djam\'{e} Seddah, Sandra K\"{u}bler, and Reut Tsarfaty. 2014.
\newblock Introducing the {SPMRL} 2014 shared task on parsing
  morphologically-rich languages.
\newblock In \emph{First Joint Workshop on Statistical Parsing of
  Morphologically Rich Languages and Syntactic Analysis of Non-Canonical
  Languages}, pages 103--109, Dublin, Ireland.

\bibitem[{Shacham and Wintner(2007)}]{ShachamWintner2007}
Danny Shacham and Shuly Wintner. 2007.
\newblock Morphological disambiguation of {H}ebrew: A case study in classifier
  combination.
\newblock In \emph{Proceedings of EMNLP/CoNLL 2007}, pages 439--447, Prague.

\bibitem[{Shao et~al.(2018)Shao, Hardmeier, and Nivre}]{ShaoHardmeierNivre2018}
Yan Shao, Christian Hardmeier, and Joakim Nivre. 2018.
\newblock Universal word segmentation: Implementation and interpretation.
\newblock \emph{Transactions of the Association for Computational Linguistics},
  6:421--435.

\bibitem[{Sima'an et~al.(2001)Sima'an, Itai, Winter, Altman, and
  Nativ}]{SimaanItaiWinterEtAl2001}
Khalil Sima'an, Alon Itai, Yoad Winter, Alon Altman, and Noa Nativ. 2001.
\newblock Building a tree-bank of {M}odern {H}ebrew text.
\newblock \emph{Traitment Automatique des Langues}, 42:347--380.

\bibitem[{Straka et~al.(2016)Straka, Haji\v{c}, and
  Strakov\'{a}}]{StrakaHajicStrakova2016}
Milan Straka, Jan Haji\v{c}, and Jana Strakov\'{a}. 2016.
\newblock {UDPipe}: Trainable pipeline for processing {CoNLL-U} files
  performing tokenization, morphological analysis, {POS} tagging and parsing.
\newblock In \emph{Proceedings of LREC 2016}, pages 4290--4297, Portorož,
  Slovenia.

\bibitem[{Yona and Wintner(2005)}]{YonaWintner2005}
Shlomo Yona and Shuly Wintner. 2005.
\newblock A finite-state morphological grammar of {H}ebrew.
\newblock In \emph{Proceedings of ACL-05 Workshop on Computational Approaches
  to Semitic Languages}, pages 9--16, Ann Arbor, MI.

\end{thebibliography}
\bibliographystyle{acl_natbib_nourl}

\end{document}